%% file: main.tex
\documentclass[10pt,twocolumn,letterpaper]{article}

\usepackage{wacv}              

\input{preamble}

\definecolor{wacvblue}{rgb}{0.21,0.49,0.74}
\usepackage[pagebackref,breaklinks,colorlinks,allcolors=wacvblue]{hyperref}

\def\wacvPaperID{*****} 
\def\confName{WACV}
\def\confYear{2027}

\title{Beyond Trial-and-Error: Agentic Optimization for Image-to-Video Adherence}

\author{Aman Tyagi$^{1}$ \quad Hemanth Boinpally$^{1}$ \quad Jonathan Chen$^{1}$ \quad Douglas Gebert$^{1}$ \quad Steven Hickson$^{2}$\\
$^{1}$Google Cloud \qquad $^{2}$Google DeepMind\\
{\tt\small \{amanty, brhemanth, jonchen, dgebert, shickson\}@google.com}
}

\begin{document}
\maketitle
\input{sec/0_abstract}
\input{sec/1_intro}

\input{sec/2_related_work}
\input{sec/3_method}
\input{sec/4_experiments.tex}
\input{sec/5_discussion.tex}

\input{sec/6_conclusion.tex}
{
    \small
    \bibliographystyle{ieeenat_fullname}
    \bibliography{main}
}

\clearpage
\input{sec/7_appendix.tex}
\end{document}

%% file: preamble.tex
\usepackage{makecell}
\usepackage{listings}

%% file: sec/0_abstract.tex
\begin{abstract}
Modern black-box Image-to-Video (I2V) models offer powerful capabilities in automated content creation, yet their lack of fine-grained control and reliability presents significant challenges in professional workflows. Their inherent stochasticity causes minor variations in textual prompts or hyperparameters to yield drastically different outputs, often necessitating inefficient, brute-force trial-and-error processes.
To address these limitations, we introduce the ``Agentic Self-Improvement'' framework, which reframes video synthesis into a closed-loop, goal-directed optimization. Our framework systematically navigates the generation parameter space using a novel two-stage approach. In the first stage, an iterative prompt optimization loop uses a multimodal Large Language Model (mLLM) to refine the input prompt. This refinement implements two automated evaluations: Davidsonian Scene Graph (DSG) queries for semantic adherence and Common Mistake Questions (CMQ) for artifact detection. In the second stage, we use Bayesian optimization to efficiently co-optimize stochastic seeds and CFG scales. This search is guided by a suite of quality metrics, including the novel Video-Text Adherence (VTA) score derived from the DSG and CMQ evaluations.
Our framework significantly outperforms unguided search methods: in human preference studies, videos generated via our agentic approach were strongly preferred over baseline outputs, achieving win rates up to 69\%. This work provides a practical and extensible methodology for enhancing the predictability and control of state-of-the-art video generation models, moving the field beyond speculative curiosities toward reliable, production-ready tools.
\end{abstract}

%% file: sec/1_intro.tex
\section{Introduction}
\label{sec:intro}

The advent of large-scale, black-box text- and image-to-video (I2V) models has unlocked powerful capabilities in automated content creation, promising a new frontier for digital media and storytelling \cite{cho2024sora,sun2024sora}. However, a significant gap remains between the generative potential of these systems and the practical demands of production-quality video synthesis \cite{wang2018video,chen2024videocrafter2}. For professional applications, the foremost requirement is precise adherence to creative briefs, where the generated video must faithfully reflect both a textual prompt and a given input image \cite{cheng2025vpo,opensora2,hu2022make,ji2024prompt}. The core challenge lies in the inherent stochasticity and sensitivity of current models; minor variations in hyperparameters, such as random seeds or classifier-free guidance (CFG) scales, can lead to dramatically different and often unpredictable outputs \cite{bethard2022we,shen2024rethinking}. This forces practitioners into a brute-force, trial-and-error workflow, generating a multitude of video variations in the hope that one serendipitously aligns with their intent—a process that is inefficient and computationally expensive. Bridging this gap requires moving beyond speculative generation toward a systematic framework for intelligently navigating the complex parameter space of these models.

To address these limitations, this paper introduces ``Agentic Self-Improvement'', a framework that re-architects the generation pipeline for I2V models from an open-loop, speculative process to a closed-loop, goal-directed optimization. Our framework employs a two-stage strategy that systematically targets the primary sources of variance: the textual prompt and the model's hyperparameters. First, a prompt optimization stage uses a multimodal Large Language Model (mLLM) to iteratively refine the user's initial prompt. To guide this refinement, the mLLM generates and scores against a structured set of evaluations: Davidsonian Scene Graph (DSG) queries \cite{cho2023davidsonian} that probe semantic content, and Common Mistake Questions (CMQ) designed to detect typical generation artifacts. Second, a hyperparameter optimization stage uses Bayesian optimization to systematically search for optimal random seeds and CFG scales. The entire process is guided by a suite of quality metrics and the novel Video-Text Adherence (VTA) score derived directly from the DSG and CMQ evaluations, providing a quantitative measure of the output's quality and alignment. The result is video content that is not only of high quality but is also aligned with the user's specified intent.

Our work makes the following primary contributions:
\begin{itemize}

\item An ``Agentic Self-Improvement'' framework for systematic, goal-directed control of black-box I2V generation models.

\item A prompt optimization method that uses an mLLM with automatically generated DSG and CMQ evaluations for iterative refinement.

\item A Bayesian optimization stage to efficiently co-optimize stochastic seeds and CFG scales.

\item A novel Video-Text Adherence (VTA) metric derived from structured DSG and CMQ evaluations to guide the optimization process and score final content.

\end{itemize}

%% file: sec/2_related_work.tex
\section{Related Work}
\label{sec:related_work}

\subsection{Advances in Text/Image-to-Video Generation}

The field of video generation has seen a paradigm shift from Generative Adversarial Networks (GANs) \cite{goodfellow2014generative} and Variational Autoencoders (VAEs) \cite{kingma2019introduction} to diffusion-based models \cite{ho2020denoising}. Early methods often struggled with temporal coherence and were limited to low-resolution outputs \cite{vondrick2016generating}. The advent of diffusion models, particularly with the introduction of large-scale models such as Veo\footnote{\url{https://deepmind.google/models/veo/}}, has marked a significant leap in generating high-fidelity, temporally consistent video sequences from textual prompts \cite{brooks2024video}. These models leverage vast datasets and sophisticated architectures, such as the Diffusion Transformer (DiT), to achieve unprecedented levels of realism and semantic accuracy \cite{peebles2023scalable}. 

More recently, the field has seen a rapid convergence toward unified autoregressive models and multimodal foundation models capable of natively processing interleaved text, image, and video tokens \cite{yu2026videomar}. However, despite these improvements in raw fidelity, their black-box nature and reliance on massive, often uncurated datasets present significant challenges in terms of controllability and reproducibility \cite{zhang2023adding,hertz2022prompt}.

\subsection{Controllability in Video Synthesis}
While text-to-video models have become increasingly powerful, relying solely on a textual prompt for control is often insufficient for professional applications that demand precise outputs \cite{brooks2024video}. This has spurred research into more fine-grained control mechanisms. One line of work focuses on incorporating additional conditioning inputs, such as a reference image for appearance control, to guide the generation process \cite{chen2023control}. Other approaches leverage spatial controls like depth maps, edge maps, or human pose information to dictate the scene's geometry and character motion \cite{zhang2023adding}. A significant challenge in this domain is the disentanglement of content from motion, and achieving precise, multi-faceted control over the generated video's content, style, and dynamics remains a key area of research \cite{zhang2023adding,blattmann2023align}.

\subsection{Automated Prompt Engineering}
Prompt engineering or prompt optimization for diffusion models has moved beyond manual tuning toward automated and systematic approaches. A significant body of work has explored gradient-based methods, which treat prompt embeddings as continuous vectors that can be optimized to maximize a differentiable reward function, a technique that has been shown to effectively discover ``hard'', interpretable prompts \cite{wen2023hard}. Another prominent direction involves the use of reinforcement learning (RL), where a policy network is trained to refine prompts based on feedback from a reward model, which can be designed to reflect human preferences or other desirable image characteristics \cite{black2023training}. Furthermore, researchers have leveraged large language models (LLMs) as optimizers themselves, employing them in black-box settings to iteratively rewrite and improve prompts based on generated image outcomes, demonstrating strong performance without requiring direct access to the diffusion model's gradients \cite{he2024automated,hao2023optimizing}. In the video domain, recent methods such as VPO \cite{cheng2025vpo}, Prompt-A-Video \cite{ji2024prompt}, and RAPO \cite{gao2025devil} optimize textual prompts via offline preference-aligned LLMs or retrieval-augmented single-pass rewriting prior to synthesis, while concurrent test-time refinement frameworks such as Critique Coach Calibration \cite{gu2025ccc} and VQQA \cite{song2026vqqa} employ closed-loop MLLM critique to iteratively evolve video prompts. Concurrently, fine-grained video evaluation benchmarks such as ETVA \cite{guan2025etva} and VBench-2.0 \cite{zheng2025vbench2} have demonstrated the value of multi-stage question answering and intrinsic faithfulness checks for assessing video synthesis. Building upon these insights, our work repurposes the Davidsonian Scene Graph (DSG) evaluation methodology \cite{cho2023davidsonian}—originally introduced for fine-grained question generation and answering (QG/A) evaluation of static images—and augments it with video-specific Common Mistake Questions (CMQ) to serve both as an active, closed-loop visual critique signal in Stage 1 and as a frozen semantic adherence objective (VTA) coupled with Stage 2 Bayesian hyperparameter optimization.

\subsection{Hyperparameter Optimization and Agentic AI}
The performance of generative models is also critically dependent on a set of hyperparameters, such as the classifier-free guidance (CFG) scale and the initial random seed \cite{ho2022classifier}. The conventional method for tuning these parameters is a brute-force grid search, which can be computationally expensive and often suboptimal. To address this, more systematic optimization techniques have been explored. By building a probabilistic model of the objective function, Bayesian optimization can intelligently explore the parameter space to find optimal configurations with fewer evaluations \cite{snoek2012practical}. Our work extends this approach by co-optimizing both deterministic hyperparameters and the stochastic random seed within a unified framework.

The concept of ``Agentic AI'' represents a shift from passive, single-turn generative models to active, goal-directed systems that can autonomously plan, reason, and act to achieve a specified objective. In the context of content creation, agentic frameworks are being developed to automate complex, multi-step generative tasks \cite{yang2023mm}. These systems often employ a large language model as a central ``agent'' that can decompose a high-level goal into a sequence of smaller steps, execute those steps using various tools (such as generative models), and then evaluate the results to refine its plan \cite{yang2023mm,huang2024dialoggen}. Our ``Agentic Self-Improvement'' framework is inspired by this paradigm. The term ``Agentic'' distinguishes our approach from standard generate-evaluate pipelines by aligning with formal definitions of autonomous LLM agents (dynamic perception, reasoning, and action \cite{xi2023rise,wang2024survey}). Specifically, our system operates as a structured two-stage sequential pipeline: in the first stage, an mLLM agent autonomously constructs a prompt-specific perception rubric (DSG/CMQ), reasons over visual and structural failures, and iteratively rewrites the prompt; in the second stage, hyperparameter search over CFG scales and random seeds is delegated to a black-box Bayesian optimizer guided by the frozen rubric and perceptual quality metrics, conceptualizing video generation as a closed-loop optimization problem.


%% file: sec/3_method.tex
\section{Methodology}
\label{sec:method}

\begin{figure*}[t]
\centering
\includegraphics[width=0.84\textwidth,clip]{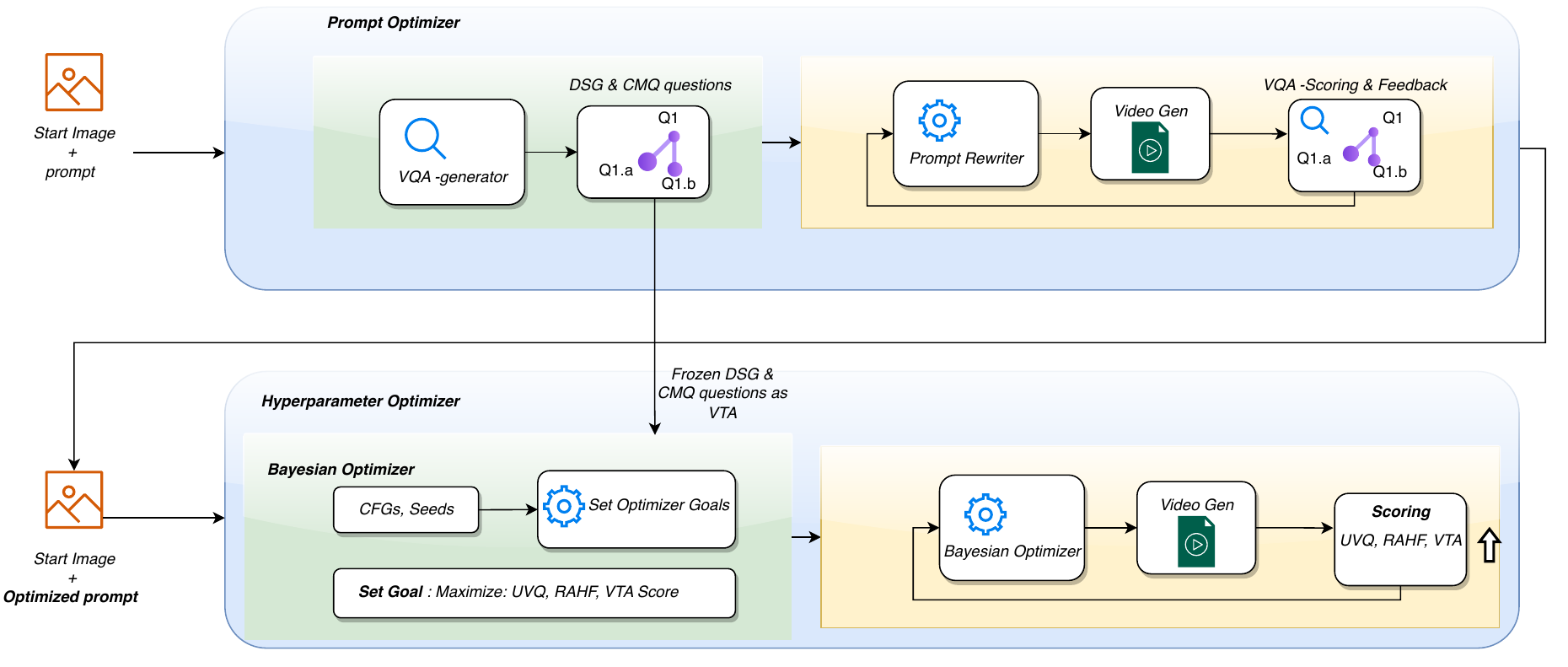} 
\caption{The two-stage architecture of our Agentic Self-Improvement framework.
Top: Prompt Optimization. A prompt is iteratively improved by generating a video and scoring its alignment against a set of auto-generated DSG and CMQ questions.
Bottom: Hyperparameter Optimization. Using the best prompt, a Bayesian search then finds optimal hyperparameters (CFGs, seeds) by maximizing a multi-objective function composed of quality metrics (UVQ, RAHF) and a Video-Text Adherence (VTA) score repurposed from the questions above.}
\label{fig1}
\end{figure*}

The Agentic Self-Improvement framework operates as a closed-loop, goal-directed system that systematically optimizes the two primary inputs to a black-box I2V model: the textual prompt and its core hyperparameters. As depicted in Figure \ref{fig1}, the framework achieves this through two main subsystems—a Prompt Optimizer module and a Hyperparameter Optimizer module—which work in concert to reliably align the final video output with the user's precise creative intent.

\subsection{Prompt Optimization via Structured Evaluation}
A primary challenge in video generation is the semantic gap between a user's prompt and the data distributions on which foundation models are trained. To bridge this gap, our framework employs an iterative prompt optimization strategy that refines the initial text prompt into an unambiguous and effective directive for the I2V model.

\begin{figure*}[t]
\centering
\includegraphics[width=0.64\textwidth,clip]{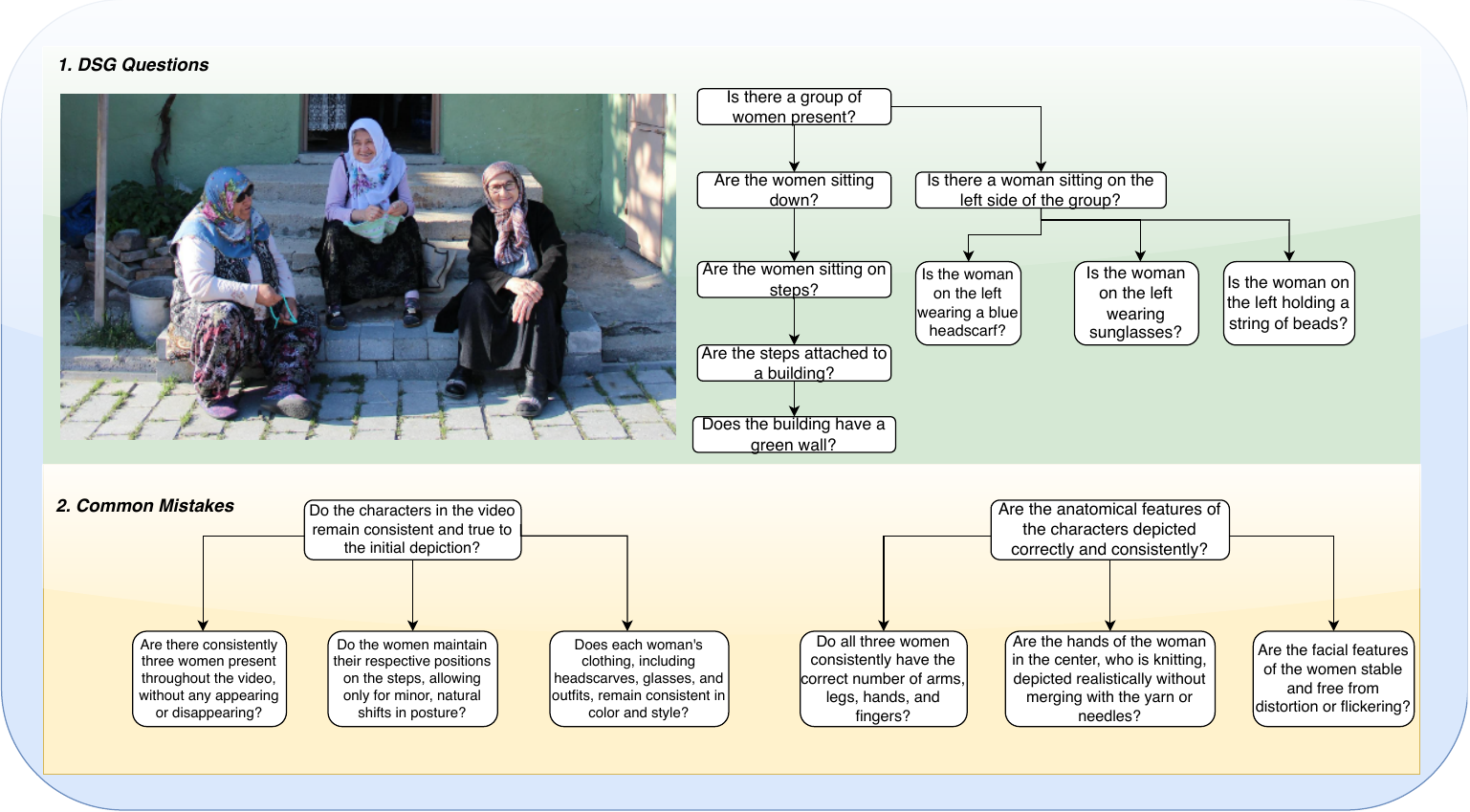}
\caption{Example of our automated question generation process for a given input image and prompt ``a group of women sitting on the steps of a building''. Top (DSG Questions): Davidsonian Scene Graph \cite{cho2023davidsonian} questions are generated to deconstruct the visual scene into semantic queries. Bottom (Common Mistakes): Common Mistake Questions are designed to target and preempt known failure modes of video generation models.}
\label{fig:dsg_cmq_example}
\end{figure*}

\subsubsection{Question Generation for Semantic Grounding}
The process begins by translating the input prompt and initial conditioning image(s) into a set of concrete, machine-verifiable questions. We use a multimodal Large Language Model (mLLM), such as Gemini, to generate two categories of evaluative queries:

Davidsonian Scene Graph (DSG) Questions: Leveraging the Davidsonian Scene Graph methodology \cite{cho2023davidsonian} for fine-grained visual evaluation, these questions probe the fundamental semantic components of the desired scene. They deconstruct the prompt into verifiable queries about the presence of agents, their actions and states, object possession, and location in a graphical format. For instance, in Figure~\ref{fig:dsg_cmq_example} (Top), the prompt ``A group of women sitting on the steps of a building'' is broken down into verifiable questions such as ``Is the group of women present?'', ``Are the women sitting down?'', and ``Are the women sitting on steps?'' This structured decomposition allows our framework to systematically check for the presence and correctness of each core semantic element.

Common Mistake Questions (CMQ): While DSG effectively covers semantic content, it may not capture common failure modes of video generation models. CMQs are designed to target these specific weaknesses, focusing on criteria such as object permanence, natural movement, and temporal consistency while also checking for visual glitches or artifacts. Continuing with the same example in Figure~\ref{fig:dsg_cmq_example} (Bottom), a CMQ could be ``Are the anatomical features of the characters depicted correctly and consistently?'' or ``Are the facial features of the women stable and free from distortion or flickering?'' These questions force the evaluation to assess the dynamic and temporal qualities crucial for plausible video synthesis.

\subsubsection{Iterative Prompt Refinement Loop}
With the structured question graphs in place, the framework initiates a self-correction loop to find the optimal prompt while holding the generator's hyperparameters (random seed and CFG scale) fixed:
\begin{itemize}

\item The I2V model generates a candidate video $V$ from the conditioning start image $I_{\text{start}}$ and the current prompt.

\item The generated video $V$ (alongside $I_{\text{start}}$ to verify fidelity to the initial depiction) is then evaluated by the mLLM Visual Question Answering (VQA) system\footnote{We use Gemini 2.5 Pro as a VQA model, which has demonstrated state-of-the-art performance on video understanding benchmarks \cite{gemini25pro-video}.}, which provides binary ``Yes/No'' answers to the full set of DSG and CMQ questions. These responses are aggregated via the hierarchical parent-masked Video-Text Adherence (VTA) score (formalized in Section~\ref{sec:eval_suite}) to serve as a quantitative alignment score.

\item Based on the score and, more specifically, the questions that were answered ``No,'' the mLLM is prompted to revise the prompt to explicitly address the identified shortcomings while preserving the original creative intent.

\item This process is repeated for a fixed number of iterations ($10$ cycles), with each step aiming to maximize the alignment score.

\item The prompt that yields the highest alignment score is passed to the next stage for hyperparameter optimization.
\end{itemize}

\subsubsection{mLLM VQA Reliability}
\label{sec:llm-vqa}
Because this prompt refinement loop relies entirely on the mLLM functioning as an automated evaluator, we must first validate its reliability against human judgment. To validate this critical component, we conducted a study to measure the agreement between the answers provided by our selected VQA model, Gemini 2.5 Pro, and a human-annotated ground truth.

For this validation, we randomly sampled 100 video-question pairs from our experiments. This sample included 50 Davidsonian Scene Graph (DSG) and 50 Common Mistake Questions (CMQ). Two human annotators, strictly calibrated on shared guidelines, each evaluated 50 disjoint video-question pairs by watching the video and providing a binary ``Yes/No'' answer to the corresponding question. These human-provided labels served as our ground truth. We then posed the same questions to the Gemini 2.5 Pro model and compared its answers to the human ground truth. The results of this analysis are presented in Table~\ref{tab:vqa_reliability}.

\begin{table}[ht]
\centering
\caption{Accuracy of Gemini 2.5 Pro on the VQA task compared to human-annotated ground truth. The high level of agreement validates its use as an automated evaluator within our framework.}
\label{tab:vqa_reliability}
\begin{tabular}{@{}lc@{}}
\toprule
\textbf{Question Type} & \textbf{Accuracy (\%)} \\ \midrule
DSG Questions        & 92\%                   \\
CMQ Questions        & 82\%                   \\ \midrule
\textbf{Overall}     & \textbf{87\%}          \\
\bottomrule
\end{tabular}
\vspace{-8pt}
\end{table}

The model was exceptionally accurate on DSG questions (92\%), which typically probe for more objective factual content (e.g., object presence). The accuracy on CMQ questions (82\%) was lower, which is expected as these questions can sometimes involve more subjective assessments of motion or temporal consistency. The 18\% CMQ error rate comprises both false positives (where fast, natural motion is occasionally misread as unnatural transitions or jump cuts) and false negatives (insensitivity to subtle temporal artifacts). To prevent reward hacking—such as yielding static videos to trivially pass temporal CMQs—our temporal coherence filters explicitly penalize 2-second freeze frames, and Stage 2 co-optimizes VTA alongside perceptual metrics (RAHF and UVQ) so VQA blind spots cannot be exploited without incurring perceptual penalties. This strong correlation with human judgment confirms that Gemini 2.5 Pro is a reliable and effective proxy for human evaluation for our specific, structured VQA task.

\subsection{Hyperparameter Optimization via Bayesian Search}
With an optimal prompt identified, the framework's second stage automates hyperparameter tuning. We employ Bayesian optimization to efficiently search for the ideal configuration of the Classifier-Free Guidance (CFG) scale, explored as a continuous variable within $[1, 15]$, and the stochastic seed, replacing inefficient manual tuning. Because random seeds are non-smooth identifiers where numerical proximity does not imply perceptual similarity, Vizier BO represents a discrete candidate set of seeds as an unordered categorical variable, modeling the joint $(\text{CFG}, \text{seed})$ space via a mixed continuous-categorical Gaussian Process (GP) Bandit kernel \cite{song2024vizier}. This mixed-space co-optimization outperforms blind random sampling through two mechanisms: (1) observations across all evaluated trials share information along the continuous CFG dimension, allowing the GP surrogate to rapidly learn the marginal effect of CFG and restrict subsequent trials to high-yield guidance regimes; and (2) across the discrete categorical seed arms, the Upper Confidence Bound (UCB) acquisition function (with default exploration constant $\sqrt{\beta}=1.8$ \cite{song2024vizier}) balances initial seed exploration against exploitation by pruning low-performing seeds whose upper confidence bounds fall below the incumbent best and reallocating evaluation budget to refine CFG scales on high-potential seeds. Vizier optimizes a multi-objective Pareto frontier over $(\text{RAHF}, \text{UVQ}, \text{VTA})$ simultaneously—from which the final Top-1 candidate is selected post-hoc—steering the generative model toward outputs with high prompt adherence and minimal visual artifacts far more efficiently than unguided grid or random search.

\subsubsection{Automated Evaluation Suite}
\label{sec:eval_suite}
The efficacy of our agentic framework hinges upon a robust suite of automated metrics that collectively form the multi-objective reward function. This suite provides a holistic assessment of each generated video across three dimensions: perceptual quality, temporal coherence, and semantic alignment.

Perceptual Quality: We employ two models to assess visual quality. First, we use the RAHF (Rich Automated Human Feedback) model \cite{liang2024rich}, a multimodal Transformer designed to simulate nuanced human quality judgments for images. For each video, our evaluation process involves sampling 10 frames at equidistant intervals. These frames are individually passed through the RAHF model—conditioned on the Stage 1 optimized prompt—to generate the overall quality score for each. To determine a single representative score for the entire video, we then select the minimum score from this set of 10 frame-level scores. Second, we use UVQ (Universal Video Quality) \cite{wang2021rich}, a CNN-based architecture for general quality prediction. For this, we use the Mean Opinion Score (MOS) derived from the average of its content, distortion, and compression networks, as described in the original work.

Temporal Coherence: We implement two metrics to evaluate temporal and motion consistency. Shot-Cut Detection \& Flow Continuity analyzes optical flow between frames to quantify motion and penalize abrupt, unintended scene changes. Concurrently, loop detection identifies and penalizes undesirable content repetition, defined as a two-second loop or two-second static freeze frame. Candidate videos triggering either detector are assigned negative penalty weights during the active Bayesian search and discarded via a hard filter prior to final Top-1 selection.

Semantic Alignment: Finally, to ensure adherence to the prompt, we employ our novel Video-Text Adherence (VTA) metric. This metric repurposes the frozen DSG and CMQ graphs generated during prompt optimization to directly score the video's content against the detailed scene expectations and text. Specifically, VTA evaluates a video by hierarchically aggregating the mLLM's binary responses ($1=\text{Yes}$, $0=\text{No}$) across the generated DSG and CMQ trees. To handle hierarchy (e.g., zeroing ``Is the dog running?'' if the parent node ``Is there a dog?'' is False), a child's score is strictly masked by its parent $p(i)$. We define the recursive node score $S_i = \mathbb{I}(\text{mLLM}(Q_i, V, I_{\text{start}}) = \text{Yes}) \cdot S_{p(i)}$, with roots defaulting to $S_{p(\text{root})}=1$. VTA is computed as the weighted average pooled over all question nodes $Q = Q_{\text{DSG}} \cup Q_{\text{CMQ}}$ across both trees: $\text{VTA}(V) = (\sum_{i \in Q} w_i)^{-1}\sum_{i \in Q} w_i S_i$. Using uniform weights ($w_i = 1$), this recursive design ensures foundational concepts (root nodes) naturally exert a mathematically higher penalty upon failure due to cascading zeroing effects on descendant nodes.

While this specific suite provides a comprehensive assessment, the framework is inherently modular. Any component can be augmented or replaced with other black-box evaluators tailored to different quality dimensions.

%% file: sec/4_experiments.tex
\section{Experiments}
\label{sec:experiments}

We validate our Agentic Self-Improvement framework through two primary experiments. First, a controlled ablation study isolates and evaluates the Prompt Optimizer module against the original V-Bench++ prompts \cite{huang2024vbench++}. Second, an end-to-end evaluation benchmarks our Bayesian search stage across the hyperparameters (seed, CFG scale) against random search baselines that approximate a typical unguided workflow. For all experiments, we ground our analysis in $N=100$ image-and-prompt-to-video tasks from V-Bench++. Crucially, our agentic framework operates as a model-agnostic wrapper that fundamentally applies to any latent diffusion video model. We utilize Veo 2.0\footnote{We use \texttt{veo-2.0-generate-001} (\url{https://developers.googleblog.com/en/veo-2-video-generation-now-generally-available/}) and generate 96 frames (4 seconds) for all videos. Although we focus our analysis on image-and-prompt-to-video generation tasks, our framework is readily adaptable to purely text-to-video workflows. While newer model families such as Veo 3.1 have been released as of early 2026, we utilize the stable, generally available Veo 2.0 API to establish a controlled baseline and ensure direct comparability across our evaluation suite.} as the experimental vehicle for this study, explicitly positioning it as a robust, representative baseline. 

\begin{figure*}[t]
\centering
\setlength{\tabcolsep}{3pt}
\renewcommand{\arraystretch}{1.1}
\begin{tabular}{@{}m{0.16\textwidth}m{0.18\textwidth}m{0.18\textwidth}m{0.18\textwidth}m{0.18\textwidth}@{}}
 & \multicolumn{1}{c}{\textbf{$t = 0.0\text{ s}$ (Start)}} & \multicolumn{1}{c}{\textbf{$t = 1.0\text{ s}$}} & \multicolumn{1}{c}{\textbf{$t = 2.0\text{ s}$}} & \multicolumn{1}{c}{\textbf{$t = 3.0\text{ s}$}} \\[2pt]
\makecell[l]{\textbf{Random}\\{\footnotesize RAHF: $0.713334$}\\{\footnotesize UVQ: $4.30$}} &
\includegraphics[width=\linewidth]{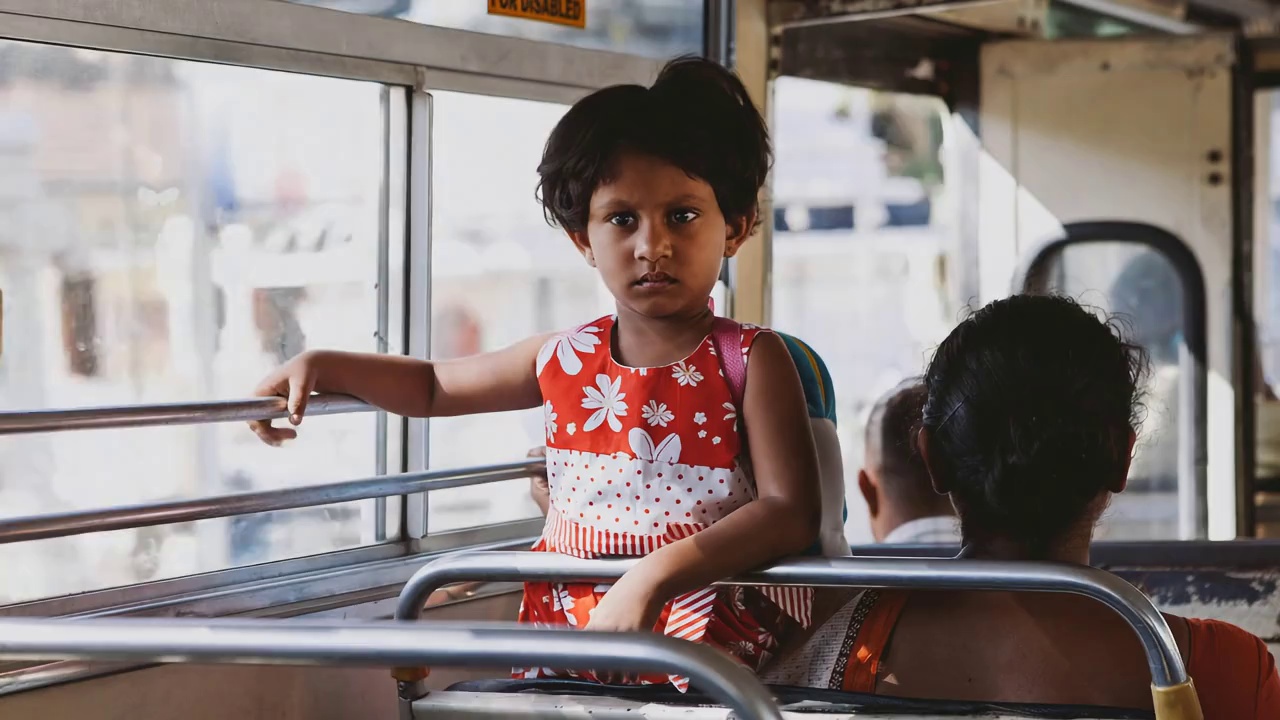} &
\includegraphics[width=\linewidth]{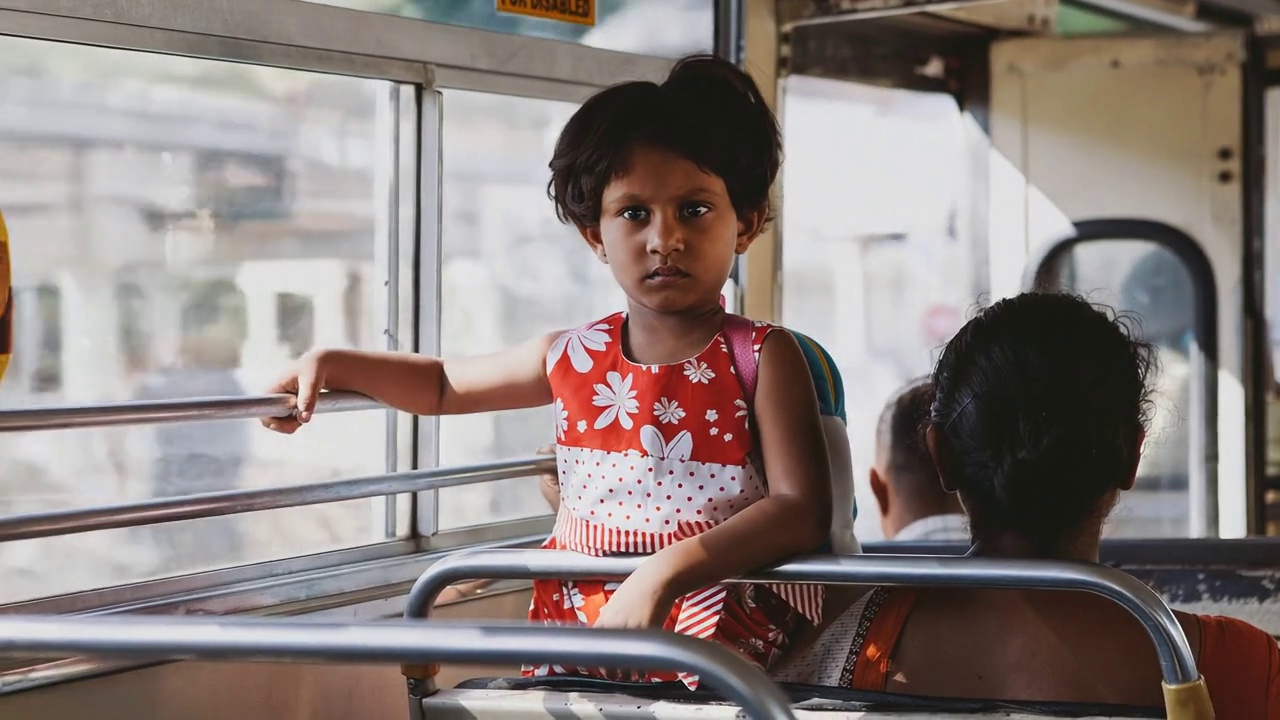} &
\includegraphics[width=\linewidth]{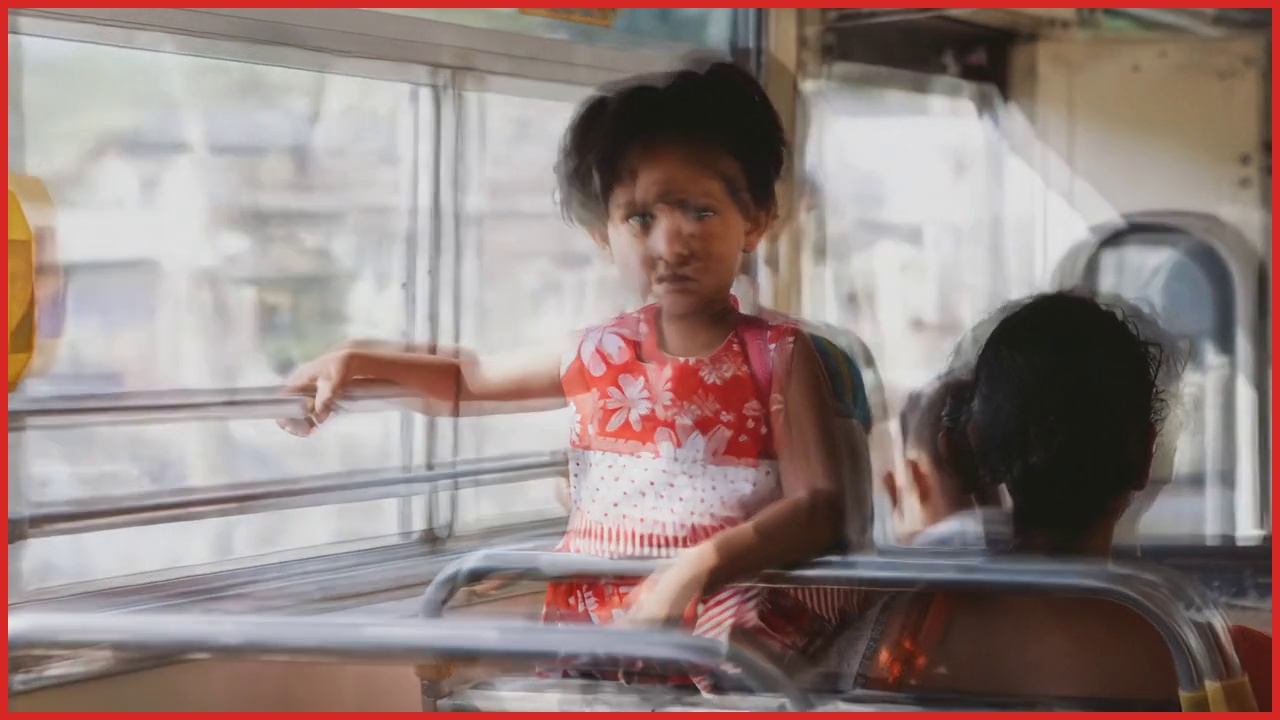} &
\includegraphics[width=\linewidth]{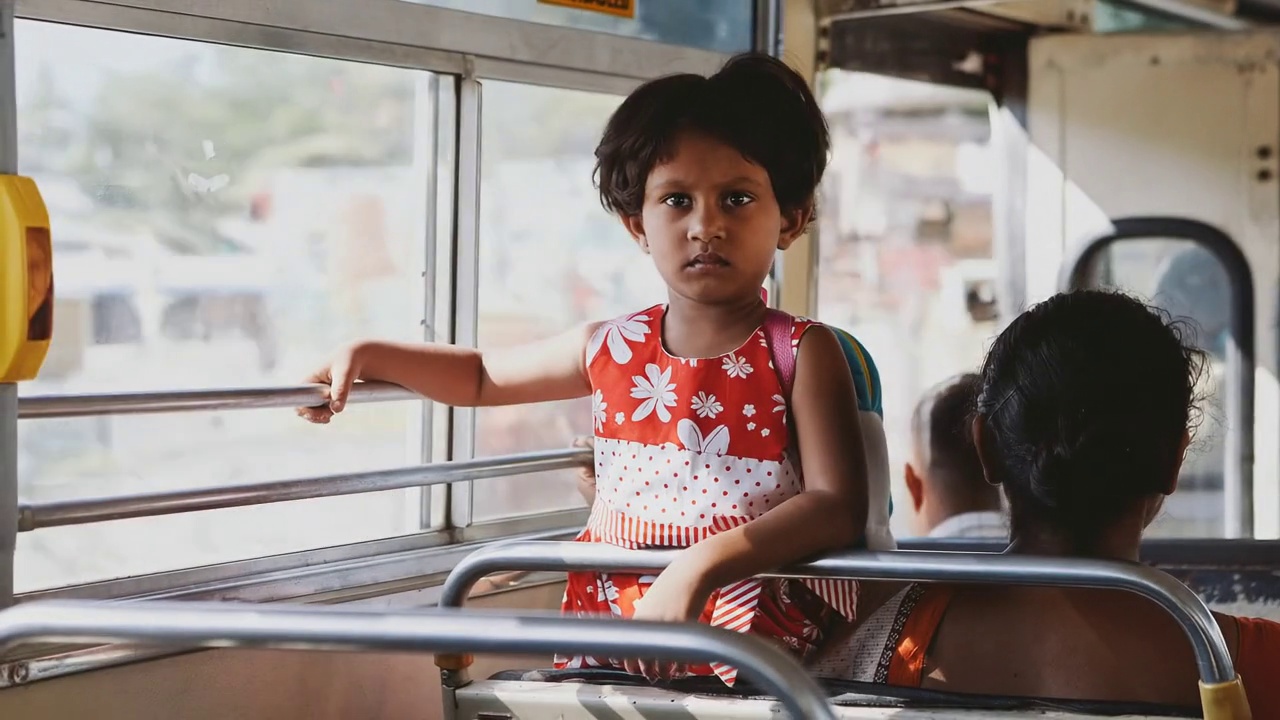} \\[4pt]
\makecell[l]{\textbf{Ours}\\{\footnotesize \textbf{RAHF: 0.911782}}\\{\footnotesize \textbf{UVQ: 4.49}}} &
\includegraphics[width=\linewidth]{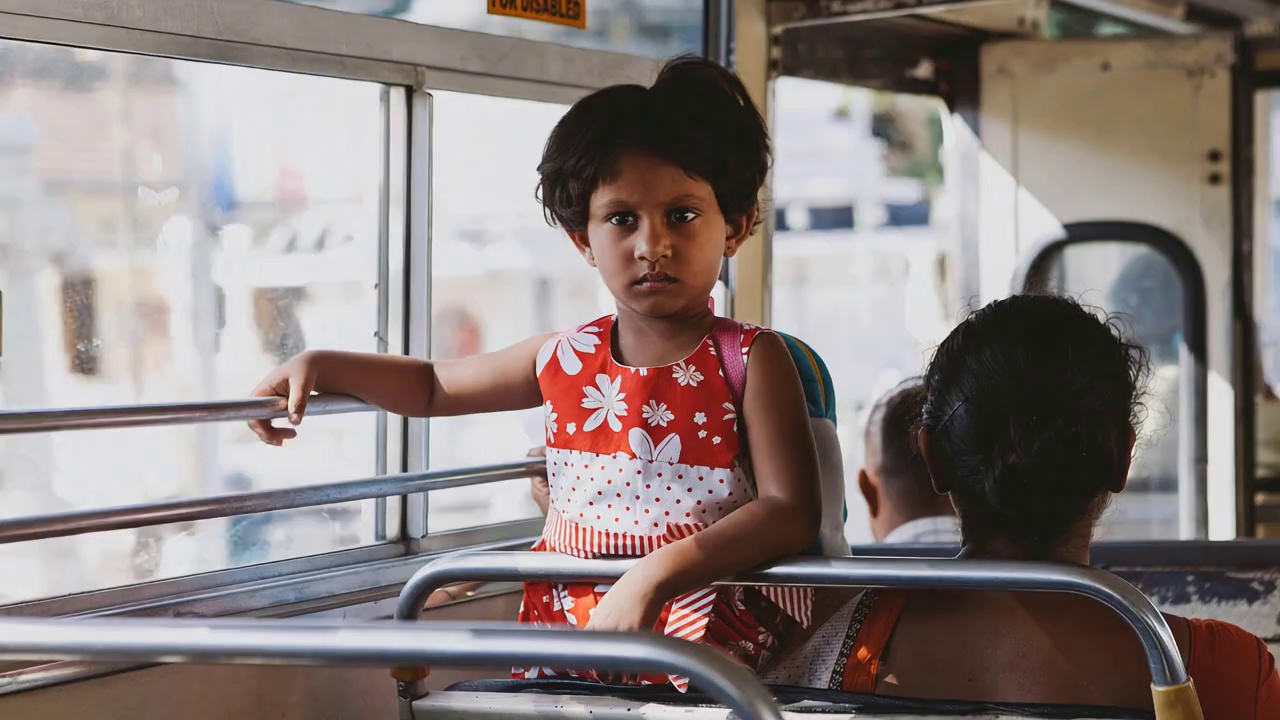} &
\includegraphics[width=\linewidth]{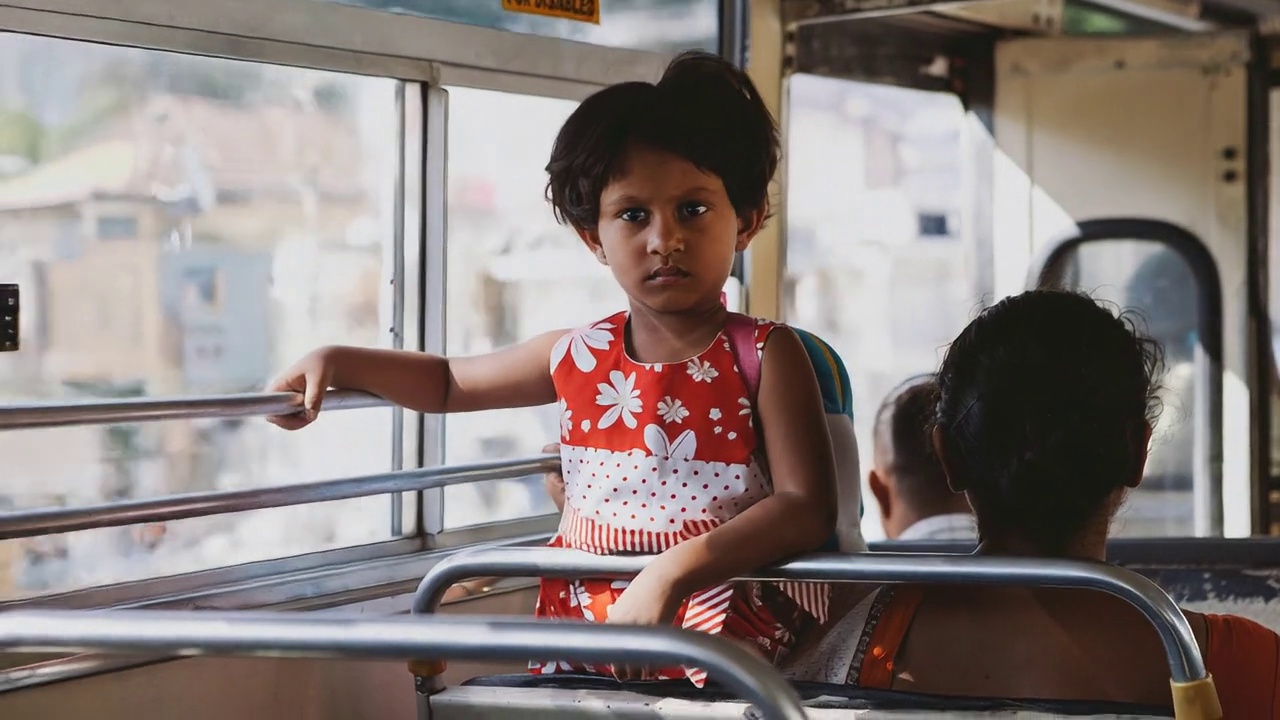} &
\includegraphics[width=\linewidth]{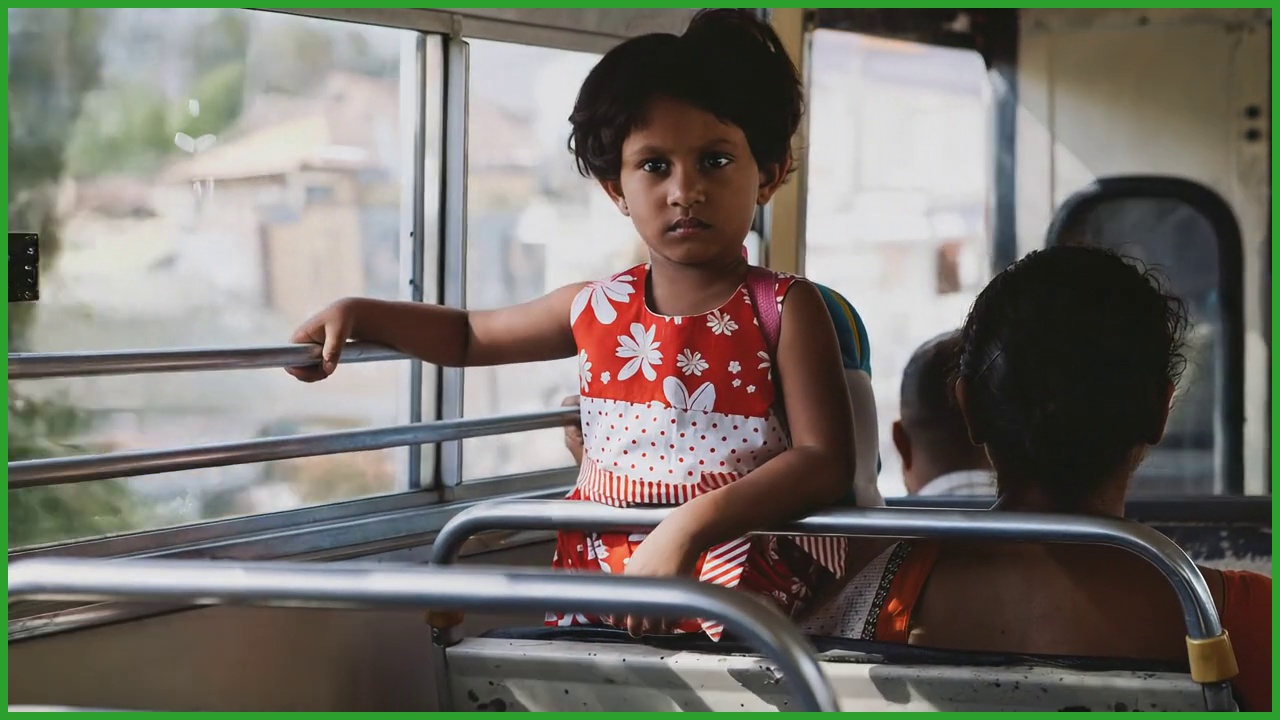} &
\includegraphics[width=\linewidth]{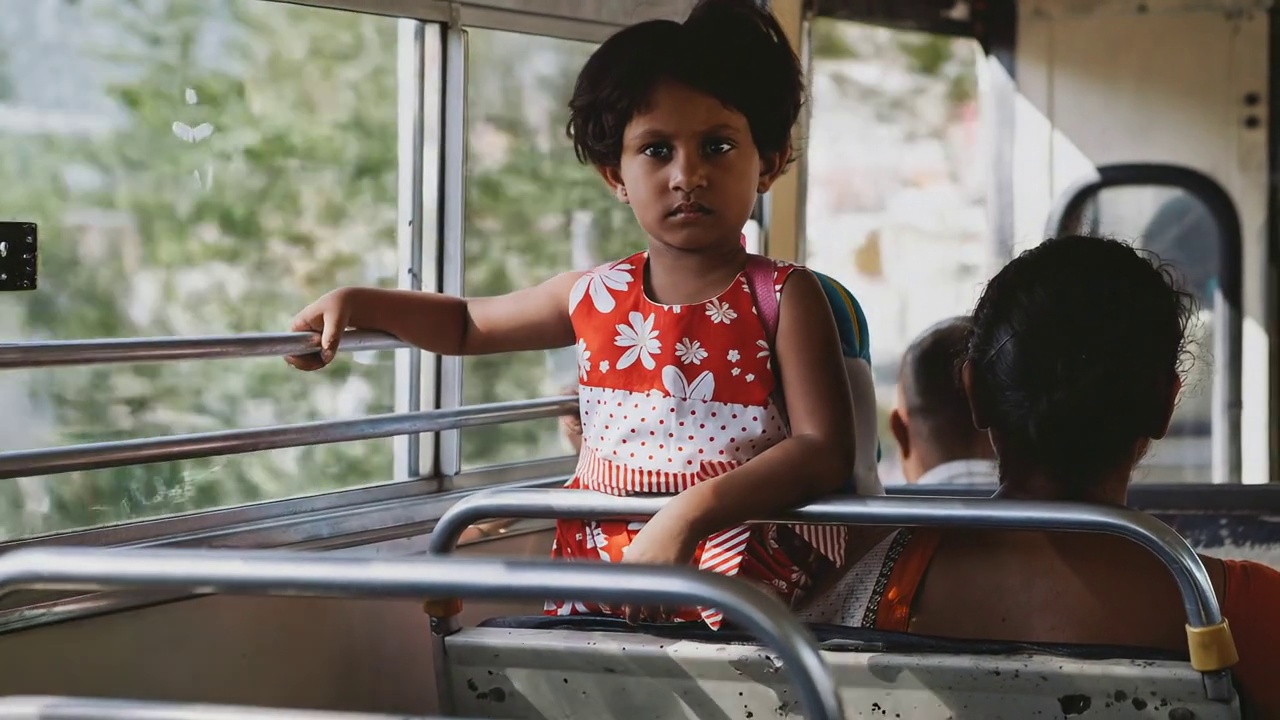} \\
\end{tabular}
\caption{Qualitative comparison between the unguided Random baseline (top; RAHF: $0.713334$, UVQ: $4.30$) and our Agentic Self-Improvement framework (bottom; RAHF: $0.911782$, UVQ: $4.49$) on a V-Bench++ task. At $t = 2.0\text{ s}$ (highlighted), the Random baseline suffers from severe motion blur and ghosting artifacts on the subject and bus interior, whereas our framework preserves sharp anatomical details, clothing patterns, and temporal consistency throughout the video.}
\label{fig:qualitative_comparison}
\vspace{-8pt}
\end{figure*}

Our primary evaluation modality is a head-to-head human preference study. We prioritize this qualitative analysis because, as we detail in the Appendix (see Section~\ref{sec:auto-eval}), standard automated metrics often lack the sensitivity to capture the nuanced improvements our framework provides. For the preference study, two expert annotators each evaluated 50 disjoint prompts ($N=100$ total) to maximize coverage, strictly calibrated on shared guidelines. To verify rater calibration, we also conducted a supplementary overlapping study on 30 video pairs, achieving 70\% observed inter-annotator agreement with a Cohen's Kappa of $\kappa=0.488$ (moderate agreement). Annotators were presented with text-image pairs alongside the corresponding videos generated by our framework and the baseline, and instructed to select the video with superior realism and overall video quality or to declare a tie. We performed two-tailed binomial sign tests on decisive (non-tied) outcomes and calculated 95\% Wilson score confidence intervals (CIs) for the unconditional win rates across all $N=100$ prompts. Across both optimization budgets ($B=10$ and $B=100$ runs per prompt), our win rates against random search are statistically significant: for $B=100$ runs, unconditional 95\% Wilson CIs range from $[50.2\%, 69.1\%]$ (UVQ; $85.7\%$ conditional non-tied win rate) to $[59.4\%, 77.2\%]$ (RAHF; $88.5\%$ conditional non-tied win rate), with $p<10^{-6}$; for $B=10$ runs, unconditional CIs range from $[32.8\%, 51.8\%]$ to $[37.5\%, 56.7\%]$ ($77.0\%\text{--}84.9\%$ conditional non-tied win rates), with $p < 3 \times 10^{-5}$. We also provide a qualitative visual comparison in Figure~\ref{fig:qualitative_comparison} to illustrate these perceptual and temporal stability improvements alongside their corresponding RAHF and UVQ scores.

\begin{table*}[ht]
\centering
\caption{Human preference evaluation comparing the top-ranked video selected by different automated metrics against a baseline across $N=100$ V-Bench++ prompts. The candidate videos were generated from budgets of $B=10$ and $B=100$ trials of Bayesian optimization, respectively. For each metric, the single best video was chosen for the head-to-head comparison. The `Avg.' metric is a simple average of normalized RAHF, VTA, and UVQ. We additionally evaluate the 100-run budget against a stronger `Best-of-Random' baseline to isolate the efficacy of the Bayesian search from the final metric selection.}
\label{tab:human_eval_metrics}
\resizebox{\textwidth}{!}{
\begin{tabular}{@{}lccccccccc@{}}
\toprule
& \multicolumn{3}{c}{\makecell{\textbf{10 Runs} \\ \textbf{(vs. Random)}}} & \multicolumn{3}{c}{\makecell{\textbf{100 Runs} \\ \textbf{(vs. Random)}}} & \multicolumn{3}{c}{\makecell{\textbf{100 Runs} \\ \textbf{(vs. Best-of-Random)}}} \\ 
\cmidrule(lr){2-4} \cmidrule(lr){5-7} \cmidrule(lr){8-10}
\makecell[l]{\textbf{Top-1 Selection} \\ \textbf{Metric}} & Baseline & \textbf{Ours} & Tie & Baseline & \textbf{Ours} & Tie & Baseline & \textbf{Ours} & Tie \\ 
& Win (\%) & \textbf{Win (\%)} & (\%) & Win (\%) & \textbf{Win (\%)} & (\%) & Win (\%) & \textbf{Win (\%)} & (\%) \\ \midrule
RAHF & 14 & \textbf{47} & 39 & 9 & \textbf{69} & 22 & 8 & \textbf{42} & 50 \\
VTA & 8 & \textbf{45} & 47 & 8 & \textbf{63} & 29 & 13 & \textbf{39} & 48 \\
UVQ & 11 & \textbf{42} & 47 & 10 & \textbf{60} & 30 & 13 & \textbf{38} & 49 \\
Avg. & 6 & \textbf{45} & 49 & 9 & \textbf{66} & 25 & 15 & \textbf{28} & 57 \\ \bottomrule
\end{tabular}
}
\vspace{-8pt}
\end{table*}

\subsection{Prompt Optimization}
To validate the impact of our Prompt Optimizer module, we conducted a controlled experiment using $N=100$ image-prompt pairs randomly sampled from V-Bench++'s image-to-video tasks. For each pair, we generated two videos while keeping all hyperparameters constant: one using the original V-Bench++ prompt and a second using the prompt generated by our optimization module after 10 iterations. These video pairs were then evaluated in a blind, head-to-head human preference study. In the comparison, videos generated from our optimized prompts were preferred by human evaluators 27\% of the time, compared to 5\% for the baseline, with the remaining 68\% resulting in a tie (two-tailed binomial sign test $p = 1.11 \times 10^{-4}$). This indicates that our method provides a statistically significant perceptual improvement in over a quarter of the cases.

\subsection{Bayesian Search}
Next, we evaluate the efficacy of our Stage 2 hyperparameter optimization against random search baselines. To strictly isolate the contribution of the Bayesian search algorithm from Stage 1 prompt refinement and ensure exact compute parity across methods, both our Bayesian optimizer and the random search baselines are conditioned on the same Stage 1 optimized prompt for each of the $N=100$ V-Bench++ image-prompt pairs. We conduct experiments under two distinct Stage 2 generation budgets: a $B=10$-generation search and a more exhaustive $B=100$-generation search. The primary baseline (`Random') randomly selects one video from the 10 (or 100) candidate videos generated via an unguided random search of hyperparameters. To further isolate the Bayesian search algorithm from the efficacy of our final metric-based candidate selection, we also evaluate the 100-generation budget against a stronger `Best-of-Random' baseline. This competitive baseline generates 100 candidates via standard random search using the Stage 1 optimized prompt, applies our temporal artifact filters, and selects the Top-1 candidate using our automated ranking metrics. Our method runs the Bayesian optimizer for the exact same number of trials ($B=10$ or $B=100$) and selects the top-ranked candidate.

To determine the single best output from our pipeline, we first apply a hard filter to discard videos with loops or shot-cuts. From the remaining pool of candidates, we test four distinct automated ranking strategies to select the Top-1 video. The selection is based on choosing the video with the highest score according to: (i) the UVQ quality model, (ii) our VTA text-adherence metric, (iii) the RAHF human-preference model, and (iv) a simple average of all three normalized scores. The human preference win rates for each of these four ranking strategies against the random baseline are presented for both the 10- and 100-generation budgets in Table~\ref{tab:human_eval_metrics}.

The results in Table~\ref{tab:human_eval_metrics} demonstrate that our agentic framework consistently produces videos that are preferred by human evaluators over the random search baseline. With a limited budget of 10 runs, our method achieves win rates of 42--47\% vs.\ 6--14\% for the baseline. The performance gap widens dramatically with a 100-generation budget, where our framework's win rate climbs to 60--69\%. This trend, coupled with a significant reduction in ties, demonstrates that our search becomes more effective at identifying perceptibly superior videos as the computational budget grows. Notably, sorting by the RAHF score yields the highest human preference (69\% win rate), suggesting it is the metric most aligned with human perception in this context.

Evaluating against the significantly stronger `Best-of-Random' baseline naturally narrows the performance gap compared to an unguided, randomly selected candidate. Because this new baseline intelligently filters and selects the best of 100 random videos, tie rates predictably increase (e.g., from 22\% to 50\% for RAHF) and our absolute win rate shifts from 69\% to 42\%. Crucially, however, our Bayesian optimizer still outperforms this competitive baseline across all individual metrics—reaching statistical significance for RAHF (42\% vs.\ 8\%, $p = 2.5 \times 10^{-5}$), VTA (39\% vs.\ 13\%, $p = 3.4 \times 10^{-4}$), and UVQ (38\% vs.\ 13\%, $p = 6.7 \times 10^{-4}$), with a positive trend for the combined Avg.\ score (28\% vs.\ 15\%, $p = 0.066$). This confirms that our framework does not merely rely on post-hoc sorting; rather, the Bayesian search actively and effectively guides the generation process into higher-quality parameter regions.

%% file: sec/5_discussion.tex
\section{Discussion}

Our experimental results validate the hypothesis of this work: a systematic, closed-loop optimization process yields perceptibly superior results compared to a standard baseline. Our experiments on prompt optimization confirm that treating the prompt as a refinable target, rather than a fixed input, is a critical step for improving semantic alignment. Because different video generation models are sensitive to distinct prompting styles \cite{gao2025devil}, this black-box optimization loop enables the framework itself to remain model-agnostic by automatically adapting the prompt to the specific target generator's sensitivities. The subsequent end-to-end evaluation via Bayesian search demonstrates that an intelligent search of the hyperparameter space is substantially more effective at discovering human-preferred videos than an unguided, random search. Collectively, these findings show that by methodically addressing the two primary sources of variance—the prompt and the hyperparameters—our framework successfully transforms video generation from a speculative, trial-and-error method into a more predictable and controllable process.

The success of the prompt optimization highlights a powerful paradigm: using the advanced multimodal understanding of mLLMs as a direct feedback signal to steer the creative output of generative models. The core principle is a ``generator-critic'' loop where the I2V model acts as the generator and the mLLM functions as a sophisticated visual critic. By generating structured DSG and CMQ questions, we provide the mLLM with a detailed rubric for its critique, forcing it to verify the fine-grained semantic and temporal properties of the generated video. This moves beyond simplistic CLIP-score similarity and instead leverages the mLLM's rich, compositional understanding of scenes, actions, and objects to guide the generation process. This approach is highly extensible, pointing toward future systems where mLLMs could provide free-form textual feedback or critique videos on more abstract qualities like narrative coherence or emotional tone, creating a powerful, self-correcting cycle for more controllable and aligned video synthesis. However, this mLLM-guided approach is naturally constrained by the current capabilities of video understanding. While state-of-the-art mLLMs excel at identifying objects and describing general scenes, their ability to comprehend nuanced motion and complex temporal dynamics is still an active area of research \cite{zhang2025q}. We anticipate that this capability will advance significantly with the increasing convergence of multimodal AI and the field of robotics \cite{team2025gemini}. The demand for embodied agents that can interpret and predict complex physical interactions from video will inevitably push the development of models with a more profound grasp of motion and causality. As these more sophisticated video-understanding mLLMs become available, they can be seamlessly integrated into our framework as more powerful ``critics,'' enabling the optimization of not just semantic content, but also physically plausible dynamics and complex character actions.

Despite its effectiveness, we acknowledge several limitations that open avenues for future research. First, regarding computational overhead, Stage 1 sequentially generates 10 videos (each evaluated via up to 40 DSG/CMQ VQA queries), followed by $B=10$ or $B=100$ candidate generations and multi-metric evaluations (VTA, 10-frame RAHF, UVQ, and optical-flow/loop filtering) in Stage 2—totaling 20 or 110 video synthesis calls per prompt. While this autonomous compute intentionally trades machine time for the tedious human trial-and-error workflows standard in professional production, future work could explore more sample-efficient optimization algorithms or distill the large mLLM critic into a smaller, specialized reward model to reduce inference latency. Second, when prompts demand complex physics that exceed the priors of the underlying generator (Veo 2.0), the Stage 1 prompt optimizer can over-constrain the text, causing Bayesian optimization to push CFG scales too high and introduce visual artifacts. Third, our framework currently operates as a decoupled two-stage process that holds the random seed and CFG constant during Stage 1; because an unfavorable fixed seed can introduce stochastic noise, Stage 1 prompt rewrites may partially adapt to seed-specific artifacts. Exploring joint or multi-seed optimization of prompts and hyperparameters simultaneously, benchmarking Stage 1 against single-pass open-loop LLM prompt rewriters \cite{cheng2025vpo,ji2024prompt,gao2025devil}, and empirically evaluating the framework across additional video generation architectures beyond Veo 2.0 remain important directions for future work.

%% file: sec/6_conclusion.tex
\section{Conclusion}
This work addressed the critical challenge of controlling black-box I2V models to achieve reliable, quality outputs that adhere to user intent. We introduced the Agentic Self-Improvement framework, a novel two-stage methodology that systematically reduces generative variance. Our approach first uses an mLLM-driven evaluation loop to optimize the text prompt for semantic clarity and then employs Bayesian optimization to discover superior hyperparameters. Through human preference studies on Veo 2.0, we demonstrate that this structured, closed-loop process significantly outperforms unguided search methods and a competitive Best-of-Random baseline. By reframing video synthesis as a goal-directed optimization problem, this work provides a practical, extensible black-box methodology for instilling predictability into I2V models, moving video generation beyond trial-and-error toward controllable production tools.

%% file: sec/7_appendix.tex
\section{Appendix}

\subsection{Automated Metric Results}
\label{sec:auto-eval}

To provide a quantitative counterpart to our human preference studies, we evaluated our generated videos using the comprehensive suite of automated metrics from the V-Bench++ benchmark \cite{huang2024vbench++}. The detailed results for each quality dimension, broken down by run configuration and Top-1 selection metric, are presented in Table~\ref{tab:automated_metrics}. While five of the six automated metrics (Aesthetic Quality, Background Consistency, Imaging Quality, Motion Smoothness, and Subject Consistency) remain tightly clustered across conditions—which we posit reflects a lack of sensitivity in standard frame- and clip-level metrics to the subtle qualitative improvements that drive human preference—Dynamic Degree exhibits a pronounced shift. Specifically, selecting candidates via VTA nearly doubles the proportion of videos with non-negligible motion (rising from $23.0\%$ in Initial Generation to $44.9\%$ in 100 Runs - VTA), as satisfying action-oriented DSG and non-static CMQ queries actively promotes motion dynamism compared to purely perceptual selectors such as UVQ ($25.5\%$).

\begin{table*}[ht]
\centering
\caption{Automated video quality metrics calculated using V-Bench++ \cite{huang2024vbench++} (mean $\pm$ standard deviation; Dynamic Degree reports the percentage of videos classified as exhibiting non-negligible motion). Five of the six metrics show minimal variation across Top-1 selection methods and optimization budgets (10 vs.\ 100 runs), corroborated by the closely clustered scores on the official V-Bench++ leaderboard \cite{VBenchLeaderboard2023}, whereas Dynamic Degree increases substantially when selecting via VTA.}
\label{tab:automated_metrics}

\resizebox{\linewidth}{!}{%
\begin{tabular}{@{}lcccccc@{}}
\toprule
\textbf{Run} & Aesthetic & Background & Dynamic & Imaging & Motion & Subject \\ 
\textbf{Configuration} & Quality & Consistency & Degree & Quality & Smoothness & Consistency \\ \midrule
\multicolumn{7}{l}{\textit{From 100 Optimization Runs}} \\
\quad 100 Runs - RAHF & $0.616\pm0.080$ & $0.961\pm0.028$ & $42.9\%$ & $69.351\pm8.460$ & $0.991\pm0.011$ & $0.948\pm0.064$ \\
\quad 100 Runs - Avg Score & $0.618\pm0.078$ & $0.965\pm0.022$ & $25.5\%$ & $70.698\pm6.939$ & $0.992\pm0.009$ & $0.963\pm0.039$ \\
\quad 100 Runs - UVQ & $0.613\pm0.080$ & $0.965\pm0.023$ & $25.5\%$ & $70.463\pm7.323$ & $0.992\pm0.009$ & $0.961\pm0.043$ \\
\quad 100 Runs - VTA & $0.616\pm0.079$ & $0.961\pm0.022$ & $44.9\%$ & $70.135\pm7.943$ & $0.991\pm0.008$ & $0.957\pm0.040$ \\
\midrule
\multicolumn{7}{l}{\textit{From 10 Optimization Runs}} \\
\quad 10 Runs - RAHF & $0.616\pm0.080$ & $0.966\pm0.024$ & $34.3\%$ & $69.334\pm8.015$ & $0.992\pm0.008$ & $0.961\pm0.041$ \\
\quad 10 Runs - Avg Score & $0.618\pm0.079$ & $0.965\pm0.021$ & $32.3\%$ & $70.350\pm7.607$ & $0.992\pm0.011$ & $0.964\pm0.039$ \\
\quad 10 Runs - UVQ & $0.617\pm0.079$ & $0.966\pm0.021$ & $30.3\%$ & $70.202\pm7.618$ & $0.991\pm0.011$ & $0.963\pm0.039$ \\
\quad 10 Runs - VTA & $0.615\pm0.080$ & $0.964\pm0.021$ & $35.4\%$ & $70.296\pm7.403$ & $0.991\pm0.007$ & $0.960\pm0.039$ \\
\midrule
\multicolumn{7}{l}{\textit{Baseline Runs}} \\
\quad Initial Generation & $0.615\pm0.081$ & $0.968\pm0.025$ & $23.0\%$ & $68.902\pm8.876$ & $0.994\pm0.004$ & $0.967\pm0.042$ \\
\quad Prompt Opt. Only & $0.615\pm0.080$ & $0.968\pm0.023$ & $27.0\%$ & $68.895\pm8.925$ & $0.994\pm0.005$ & $0.967\pm0.040$ \\
\bottomrule
\end{tabular}%
} 
\end{table*}

\subsection{Prompts for Question Generation}
\label{sec:llm_prompts}

Below are the detailed prompts provided to the mLLM for generating the structured DSG and CMQ question graphs used in our framework.

\subsubsection{Prompt for Davidsonian Scene Graph (DSG) Questions}

\begin{lstlisting}[numbers=none, breaklines=true, breakatwhitespace=true, columns=fullflexible, basicstyle=\small\ttfamily]
Analyze the provided text prompt and optional start/end frame images. Based on this information, generate a hierarchical (tree-like) set of questions following the principles of Davidsonian Scene Graphs (DSG). Identify key scene components (e.g., agents, actions, patients, locations, attributes, manners). For each component, formulate a specific yes/no question where a "yes" answer indicates correct depiction.

Organize these components and their questions into a JSON object. The JSON structure should have a main key as "scene_graph", which is a list of root elements. Each element should have at least "id" (a unique identifier for the question node), "element_type" (e.g., "agent", "action", "object", "location", "attribute"), "description" (a brief description of the element), "question" (the yes/no question), and "children" (a list of sub-elements, which can be empty).

Example of an element structure:
{
    "id": "agent_red_bird_1",
    "element_type": "agent",
    "description": "A red bird",
    "question": "Is there a red bird present?",
    "children": [
        {
            "id": "action_bird_flying_1",
            "element_type": "action",
            "description": "The red bird is flying",
            "question": "Is the red bird flying?",
            "children": []
        }
    ]
}

Ensure the entire output is a single valid JSON object. Ensure that there are no more than 20 total questions.

Initial Prompt: "[User's Initial Prompt]"
\end{lstlisting}

\subsubsection{Prompt for Common Mistake Questions (CMQ)}

\begin{lstlisting}[numbers=none, breaklines=true, breakatwhitespace=true, columns=fullflexible, basicstyle=\small\ttfamily]
Analyze the provided text prompt and optional start/end frame images. Based on this information, generate a hierarchical (tree-like) set of questions targeting potential common video generation mistakes that might occur when trying to generate a video for this prompt.

Consider only the issues such as:
    - Unnatural Transitions
    - Anatomical Inconsistencies (missing/extra limbs/digits, distorted parts)
    - Environmental Inconsistencies (objects appearing/disappearing)
    - Illogical Changes within shots
    - Static Elements that should move
    - Extra characters and objects
    - Human characters speaking (they should never speak unless specified explicitly)
    - Inconsistent Detail/Texture

Organize these components and their questions into a JSON object. The JSON structure should have a main key as "common_mistakes_graph", which is a list of root elements. Each element should have at least "id" (a unique identifier for the question node), "element_type" (e.g., "mistake_category", "specific_mistake"), "description" (a brief description of the mistake/category), "question" (the yes/no question where "yes" indicates no mistake), and "children" (a list of sub-elements, which can be empty).

Example of an element structure:
{
    "id": "mistake_cat_transitions_1",
    "element_type": "mistake_category",
    "description": "Unnatural Transitions",
    "question": "Are all transitions between shots smooth and logical?",
    "children": [
        {
            "id": "specific_mistake_jump_cut_1",
            "element_type": "specific_mistake",
            "description": "Potential jump cut between scene A and B",
            "question": "Is the transition from scene A to B free of jarring jump cuts?",
            "children": []
        }
    ]
}

Ensure the entire output is a single, complete, and valid JSON object. Make sure all brackets and braces are correctly matched and closed. Ensure that there are no more than 20 questions in the entire tree. All questions should be framed as a yes or no question such that "yes" is the correct answer (i.e., the mistake is avoided) and a "no" answer indicates a common mistake might be present.

Initial Prompt: "[User's Initial Prompt]"
\end{lstlisting}